# Subspace clustering based on low rank representation and weighted nuclear norm minimization

**Song Yu[a], Wu Yiquan[a, b, c, d]**

[a] Nanjing University of Aeronautics and Astronautics, School of Electronic and Information Engineering, Jiangjun Avenue No. 29, Nanjing, China, 211106

[b] Key Laboratory of Port, Waterway and Sedimentation Engineering of the Ministry of Transport, Nanjing Hydraulic Research Institute, Guangzhou Street No. 223, Nanjing, China, 210029

[c] State Key Laboratory of Urban Water Resource and Environment, Nangang District, Huanghe Street No. 73, Harbin, China, 150090

[d] The Key Laboratory of Rivers and Lakes Governance and Flood Protection of Yangtse River Water Conservancy Committee, Huangpu Street No. 23, Wuhan, China, 430010

**Corresponding author:** Song Yu, e-mail: 519559374@qq.com, postal address: Nanjing University of Aeronautics and Astronautics, School of Electronic and Information Engineering, Jiangjun Avenue No. 29, Nanjing, China, 211106, telephone: 86 13952027101

**Abstract:** Subspace clustering refers to the problem of segmenting a set of data points approximately drawn from a union of multiple linear subspaces. Aiming at the subspace clustering problem, various subspace clustering algorithms have been proposed and low rank representation based subspace clustering is a very promising and efficient subspace clustering algorithm. Low rank representation method seeks the lowest rank representation among all the candidates that can represent the data points as linear combinations of the bases in a given dictionary. Nuclear norm minimization is adopted to minimize the rank of the representation matrix. However, nuclear norm is not a very good approximation of the rank of a matrix and the representation matrix thus obtained can be of high rank which will affect the final clustering accuracy. Weighted nuclear norm (WNN) is a better approximation of the rank of a matrix and WNN is adopted in this paper to describe the rank of the representation matrix. The convex program is solved via conventional alternation direction method of multipliers (ADMM) and linearized alternating direction method of multipliers (LADMM) and they are respectively refer to as WNNM-LRR and WNNM-LRR(L). Experimental results show that, compared with low rank representation method and several other state-of-the-art subspace clustering methods, WNNM-LRR and WNNM-LRR(L) can get higher clustering accuracy.

## 1. Introduction

In the practice of image processing and machine learning, one often has to deal with high-dimensional data samples. These high-dimensional data samples will greatly increase the computational complexity and memory requirements of algorithms. They also severely affect the performance of the algorithms due to the insufficient number of samples with respect to the ambient space dimension and noise effect [1], [36]. Fortunately, in practice data are not unstructured. High-dimensional data often contains some type of structure that enables efficient representation and processing [39]. These data samples often lie in low-dimensional structures. The low-dimensional structures where the data samples lie can be well modeled by linear low-dimensional subspaces embedded in a high-dimensional ambient space. Several types of visual data, such as tracked feature point trajectories of a moving object in a sequence of video frames [2], face images of a subject under fixed pose and varying lighting conditions [3] and multiple instances of a hand-written digit with different translations and rotations [4] lie in a low-dimensional subspace of the ambient space [36]. Consequently, the collection of data samples drawn from multiple classes or categories lies in a union of low-dimensional subspaces [36], [39]. Subspace clustering refers to the problem of clustering a set of data samples drawn from multiple linear subspaces to their respective subspaces. Subspace clustering has found wide applications in computer vision [5], [6], [7], image processing [8], [9], pattern recognition, temporal video segmentation [10] and system identification [11]. In dealing with real visual data such as tracked feature point trajectories of several moving objects in a sequence of video frames, face images of several subjects under fixed pose and varying lighting conditions and multiple instances of several hand-written digits with different translations and rotations, subspace clustering methods have demonstrated promising performance.

*1.1 Related work in subspace clustering*

A thorough investigation of existing subspace clustering methods can be found in [12]. Here we briefly summarize the related work in subspace clustering. Existing algorithms can be divided into four main categories: iterative, algebraic, statistical, and spectral clustering-based methods [12], [36].

**Iterative methods.** Typical iterative approaches include K-subspaces [13], [14] and median K-flats [15]. There are two main steps used in iterative approaches. The first step is assigning data samples to subspaces and the second step is fitting a subspace to each cluster. The two steps are then iterated until convergence [12]. These approaches are not only very sensitive to initialization but can also easily get stuck in local minimum. Besides, they generally require knowing the dimensions and number of the subspaces [12]. Iterative statistical approaches, as one kind of iterative methods, also contain these two steps. Typical iterative statistical approaches include mixtures of probabilistic PCA (MPPCA) [16] and multistage learning (MSL) [17], [18]. These algorithms assume that the distribution of the data samples inside each subspace obeys Gaussian distribution and alternate between the above two steps by applying expectation maximization [12]. They also suffer from the drawbacks of general iterative methods.

**Algebraic methods.** Factorization-based algebraic approaches [19], [20], [21] seek to approximate the given data matrix as a product of two matrices such that the support pattern for one of the factors provides the segmentation of the samples [39]. These approaches can correctly cluster the data samples when the subspaces are independent, but may fail when this independent condition is violated [12]. Moreover, they are sensitive to noise and outliers in the data. In order to deal with noise and outliers, extra regularization terms are added to modify the formulations of these methods. Such modifications usually lead to non-convex optimization problems which need heuristic algorithms to solve [39]. One of the factorization-based approaches is called the shape interaction matrix method [19]. The data matrix is factorized by using singular value decomposition and the shape interaction matrix is built using the first $r$ right singular vectors where $r$ is the estimated rank of the data matrix. The shape interaction matrix can be seen as a similarity matrix between the data samples in the data matrix and spectral clustering can be applied using this matrix [12]. The construction of the shape interaction matrix requires knowledge of the rank of the data matrix and using the wrong rank can lead to very poor results [12]. Another kind of algebraic approaches use geometric structure of the data samples. Typical approaches of this kind include generalized principal component analysis (GPCA) [10], [22]. The GPCA approach first fits the data samples with a set of polynomials. The gradients of all the polynomials at a point give a basis for the orthogonal complement of the subspace containing that point from which a basis of the subspace containing that point can also be obtained. Then subspace segmentation is done by assigning data samples to these subspaces with known basis [12]. However, there are several drawbacks of this method. First, this method is sensitive to noise and it is difficult to estimate the polynomials when the data are contaminated by noise. Second, its complexity increases exponentially in terms of the number and dimensions of subspaces so it is applicable only when the data dimension is low and the number of subspaces is small [12].

**Statistical methods.** Statistical approaches can be further divided into several kinds such as iterative statistical approaches, robust statistical approaches and information-theoretic statistical approaches [12]. Iterative statistical approaches can also be seen as iterative methods and have been discussed. The agglomerative lossy compression (ALC) [5] algorithm, as one of the information-theoretic statistical approaches, assumes that the data are drawn from a mixture of degenerate Gaussians. ALC looks for the segmentation of the data that minimizes the coding length needed to fit the points with a mixture of degenerate Gaussians up to a given distortion [12]. The minimization over all possible segmentation of the data samples is NP-hard. ALC deals with this issue by using an agglomerative clustering method [12]. Each data sample is considered as a separate group initially. At subsequent iterations, two groups are merged as one group if merging causes the maximum decrease of the coding length [12]. The algorithm stops when the coding length cannot be further decreased. ALC can handle noise and outliers in the data naturally. Also it does not need to know the number and dimensions of the subspaces. However, the number of subspaces found by the algorithms is closely related to the distortion parameter. In addition, there is no theoretical proof for the optimality of the agglomerative clustering method [12]. Random sample consensus (RANSAC) [9] is one of the robust statistical approaches and it finds the subspaces in a greedy fashion by fitting one subspace at a time [12]. The method recovers a basis for the first subspace along with the set of inliers and then the inliers are removed from the current data set. The step is then repeated to find the second subspace and so on until all the subspaces are recovered [12]. RANSAC can deal with noise and outliers explicitly. Besides, RANSAC does not need to know the number of subspaces beforehand. However, the dimensions of the subspaces must be known and equal. Moreover, the complexity of the algorithm increases exponentially in the number and dimension of the subspaces [12].

**Spectral clustering-based methods.** Spectral clustering-based approaches first build a similarity matrix between the data samples and then uses spectral clustering [23], [24] to this similarity matrix to obtain the clustering of the data [12]. The spectral clustering-based approaches can be divided into two kinds which are local spectral clustering-based approaches and global spectral clustering-based approaches based on the information used when constructing the similarity matrix [12]. Local spectral clustering-based approaches such as local subspace affinity (LSA) [25], locally linear manifold clustering (LMMC) [26] and spectral local best-fit flats (SLBF) [27], [28] use local information around each point to build a similarity between pairs of points [12]. The underlying assumption is that the nearest neighbors of a data point are from the same subspace as the data point and these data points can be used to estimate the local linear subspace of that data point. However, this assumption is generally not satisfied since nearby points can be from different subspaces. For example, points near the intersection of two subspaces are from different subspaces but they are close to each other. In addition, they are sensitive to the right choice of the neighborhood size to compute the local information at each point [12].

Global spectral clustering-based approaches try to overcome these difficulties by building more reasonable similarities between data points using global information. Spectral curvature clustering (SCC) [29] uses the concept of polar curvature to define multiway affinities between data points within an affine subspace [12]. SCC can deal with noise in the data. However, SCC requires knowing the number and dimensions of subspaces and that the dimension of subspaces to be equal. In addition, the complexity of building the multiway affinities between data points grows exponentially with the dimensions of the subspace; hence, in practice, a sampling strategy must be used to reduce the computational cost [12]. The recent advances in sparse [30], [31], [32] and low-rank [33], [34], [35] recovery algorithms lead to several promising global spectral clustering-based approaches such as sparse subspace clustering (SSC) [36], [37], [38], low-rank representation (LRR) [39], [40], [41] and low-rank subspace clustering (LRSC) [42], [43]. These algorithms aim to find a representation matrix of the data samples using a dictionary composed of the data itself. Usually a convex optimization problem is formed and the representation matrix can be solved [12]. The representation matrix is then used to build a similarity matrix from which the segmentation of the data is obtained. The advantages of these methods with respect to most state-of-the-art algorithms are that they can handle noise and outliers in data explicitly. Moreover, they do not need to know the dimensions and the number of subspaces a priori [12].

*1.2 The motivation of the paper*

First we briefly introduce the algorithmic procedure of the low rank representation based subspace clustering method. The standard low rank representation based subspace clustering method [39] consists of solving the following convex optimization problem:

$$\min_{Z,E} \|Z\|_* + \lambda \|E\|_{2,1}, \quad \text{s.t.} \quad X = AZ + E \tag{1}$$

Where $X \in \mathbb{R}^{m \times n}$ is the data matrix and can be written as $X = [x_1, x_2, ..., x_n]$ where every column of the data matrix is a data sample which is *m*-dimensional and there are totally *n* data samples in the data matrix, $A$ is a dictionary and $A$ is usually chosen as the matrix $X$ for convenience, $Z$ is the representation matrix which represents the data samples as linear combinations of the data samples, $E$ is an error matrix and $\lambda$ is a weighting parameter. In the above convex optimization problem, the sum of the nuclear norm of the representation matrix and the $l_{2,1}$ norm of the error matrix weighted by the parameter $\lambda$ is to be minimized and the convex optimization problem can be efficiently solved via ADMM [44], [45]. After obtaining the representation matrix, an affinity matrix is constructed from the singular value decomposition of the representation matrix. Spectral clustering [23], [24], [46] is then applied to the affinity matrix to obtain the final clustering results. The reason nuclear norm minimization is to be achieved for the representation matrix is that we want the representation matrix to be of a low rank matrix and nuclear norm is a convex relaxation of the rank of a matrix. The rank of the representation matrix obtained is closely related to the final clustering accuracy. Since the representation matrix is expected to be a low rank matrix, we can reasonably image that the clustering accuracy will decrease when the rank of the representation matrix is increasing.

In a recently published work [47] considering low rank approximation, the authors propose to minimize the partial sum of singular values instead of minimizing the nuclear norm. The work [48] also considers to minimize the partial sum of singular values. The reason for using partial sum is that the priori target rank information can be incorporated in the minimization procedure. The objective function of the partial sum of singular values can be written as:

$$\min_{A_{(p)}, E_{(p)}} \|A_{(p)}\|_{p=N_{(p)}} + \lambda_{(p)} \|E_{(p)}\|_1, \quad \text{s.t.} \quad O_{(p)} = A_{(p)} + E_{(p)} \tag{2}$$

Where $\|A_{(p)}\|_p = \sum_{i=p+1}^{\min(m_{(p)}, n_{(p)})} \sigma_i(A_{(p)})$ and $N_{(p)}$ is the target rank of $A_{(p)}$ which can be derived from the problem definition. $O_{(p)} \in \mathbb{R}^{m_{(p)} \times n_{(p)}}$ is the observation matrix and the subscript (p) is used for distinguishing the variables used in the objective function in partial sum from the variables previously used in the object function in low rank representation. Eq. (2) minimizes the rank of residual errors of $A_{(p)}$ against the target rank, instead of the nuclear norm. Using the technique from the partial sum of singular values, we can control the rank of the matrix obtained since the matrix obtained by solving the optimization problem (2) has the property that $\text{rank}(A_{(p)}) \geq N_{(p)}$. This technique provides us a convenient tool for investigating the relationship between the rank of the representation matrix and the final clustering accuracy. We can modify the objective function of the original low rank representation based subspace clustering by minimizing the partial sum of singular values of the representation matrix. The modified objective function can be written as:

$$\min_{Z,E} \|Z\|_{p=N} + \lambda \|E\|_{2,1}, \quad \text{s.t.} \quad X = AZ + E \tag{3}$$

By setting *N* to different values, we can obtain the representation matrices with different ranks. To clearly show the relationship between the rank of the representation matrix and the clustering accuracy, we have done a set of experiments using Extended Yale B dataset [49]. We choose 10 individual subjects from the dataset and each subject contains 64 images acquired under different illumination conditions. The data matrix is of size $2016 \times 640$ and each column of the data matrix is a vector from transforming the face image. Table 1 shows the clustering accuracy when setting *N* to different values. From the table we can conclude that the accuracy will decrease when the rank of the representation matrix increases. This fact naturally raises the following question: can we obtain a lower rank representation matrix than nuclear norm minimization? If the answer to the above question is positive, we can expect the lower rank representation matrix will lead to a higher clustering accuracy than the original nuclear norm minimization based low rank representation.

Table 1 The relationship between the rank of the representation matrix and clustering accuracy

|  | Rank | Accuracy |  | Rank | Accuracy |
|---|---|---|---|---|---|
| *N*=0 | 144 | 96.72 | *N*=65 | 171 | 72.66 |
| *N*=5 | 145 | 95.63 | *N*=70 | 174 | 71.41 |
| *N*=10 | 145 | 95.63 | *N*=75 | 174 | 75.00 |
| *N*=15 | 146 | 95.63 | *N*=80 | 178 | 59.22 |
| *N*=20 | 148 | 95.47 | *N*=85 | 181 | 59.38 |
| *N*=25 | 150 | 96.41 | *N*=90 | 185 | 58.59 |
| *N*=30 | 152 | 95.31 | *N*=95 | 190 | 59.22 |
| *N*=35 | 155 | 95.31 | *N*=100 | 195 | 57.34 |
| *N*=40 | 158 | 95.00 | *N*=110 | 201 | 52.19 |
| *N*=45 | 161 | 94.38 | *N*=120 | 205 | 49.22 |
| *N*=50 | 164 | 94.38 | *N*=130 | 207 | 40.16 |
| *N*=55 | 167 | 92.34 | *N*=140 | 210 | 10.31 |
| *N*=60 | 169 | 75.94 | *N*=150 | 215 | 10.78 |

As pointed out by [50], [51], nuclear norm minimization regularizes each singular value equally and this will restrict its capability and flexibility in dealing with many practical problems. When dealing with low level vision problems such as image denoising, [50], [51] suggest using weighted nuclear norm minimization (WNNM) to regularize the data matrix. WNNM based image denoising algorithm achieves the best image denoising performance so far and this fact reveals that WNN can better deal with low rank matrix approximation problem when dealing with image denoising problem. Since the singular values of the data matrix formed by the similar patches from the noisy image has clear physical meanings, weights are formed in a non-descending order which penalizes less the larger singular values. With a similar idea, Chen et al [52] and Gaiffas and Lecue [53] also propose to use an adaptive weighted nuclear norm and the weights used are also arranged in a non-descending order. Inspired by the ideas from the above work, we may use the WNNM to regularize the representation matrix in order to get a lower rank matrix. In image denoising problem, larger singular values should receive lesser penalty and smaller singular values should receive heavier penalty since larger singular values of the data matrix quantify the information of its underlying principal directions. Similarly, in multivariate linear regression problem, larger singular values should be penalized less to help reducing the bias and smaller singular values should be penalized heavily to help promoting sparsity. However, in this paper we propose the opposite arrangement method. In the proposed WNNM based low rank representation method, larger singular values should receive heavier penalty while smaller singular values should receive lesser penalty in order to obtain a lower rank representation matrix. The weights used in this paper is arranged in a non-ascending order and consequently the WNNM based low rank representation is still a convex optimization problem [50] which could be solved via ADMM or LADMM efficiently.

The rest of the paper is organized as follows. Section 2 introduces the concept of WNN and applies the WNN to the low rank representation based subspace clustering method. The solution of WNNM based low rank representation model using ADMM is also presented in section 2. Section 3 proposes to use LADMM to solve the WNNM based low rank representation model. Experimental results regarding the comparison of the proposed algorithm and several other state-of-the-art subspace clustering algorithms are shown in section 4. Section 5 presents some conclusions and points out some possible future research directions.

## 2. WNN and WNN applied to the low rank representation based subspace clustering method

*2.1 The definition of WNN and its relationship with previous low rank approximation methods*

**Principal component analysis (PCA).** The goal of PCA is to find a low-rank approximation of a given data matrix and its mathematical formulation can be written as:

$$\min_{A} \| D - A \|_F^2 \text{ subject to } \text{rank}(A) \leq r \tag{4}$$

where $D$ is a noise corrupted data matrix and $D = A + E$, where $A$ is an unknown low-rank matrix and $E$ represents the noise. The optimal solution to this problem is given by $A = U_1 \Sigma_1 V_1^T$, where $U_1$, $\Sigma_1$ and $V_1$ are obtained from the top $r$ singular values and singular vectors of the data matrix $D$.

**Nuclear norm minimization (NNM) based low rank approximation.** When $r$ is unknown, the problem of finding a low-rank approximation can be written as:

$$\min_{A} \text{rank}(A) + \frac{\alpha}{2} \| D - A \|_F^2 \tag{5}$$

where $\alpha > 0$ is a parameter. Since this problem is in general NP hard, a common practice is to replace the rank of $A$ by its nuclear norm $\| A \|_*$, i.e., the sum of its singular values, which leads to the following convex problem:

$$\min_{A} \| A \|_* + \frac{\alpha}{2} \| D - A \|_F^2 \tag{6}$$

Equation (6) is also referred to as the nuclear norm proximal (NNP) operator with $F$-norm data fidelity. The optimal solution to this problem can be efficiently obtained by the singular value thresholding (SVT) operator [54]. It is given by $A = U \mathcal{S}_{\frac{1}{\alpha}}(\Sigma) V^T$, where $D = U \Sigma V^T$ is the SVD of $D$ and $\mathcal{S}_{\frac{1}{\alpha}}(\Sigma)$ is the soft thresholding operator

$$\mathcal{S}_\varepsilon(x) = \begin{cases} x - \varepsilon & x > \varepsilon \\ x + \varepsilon & x < -\varepsilon \\ 0 & \text{else} \end{cases} \tag{7}$$

The difference between PCA based low rank approximation with the NNM based low rank approximation is that the former solution performs hard threholding on the singular values of $D$ while the latter solution performs soft thresholding on the singular values of $D$. Notice that the singular values of $D$ in the NNM based low rank approximation is soft thresholded by the same number. The underlying assumption is that the importance of different singular values of $D$ are the same so they should be soft thresholded by the same amount. This assumption is definitely not correct in practice so WNN is proposed which gives different weights to different singular values of $D$.

**Weighted nuclear norm minimization (WNNM) based low rank approximation.** WNN is first proposed in the literature [52] for multivariate linear regression applications. The WNN of a matrix $X$ is defined as:

$$\| X \|_{w,*} = \sum_i w_i \sigma_i(X) \tag{8}$$

Where $w = [w_1, ..., w_n]$ and $w_i \geq 0$ is a non-negative weight assigned to $\sigma_i(X)$. The WNN can be seen as a generalization of the nuclear norm where the weights are all the same and equal to one. The WNN can improve the flexibility of the nuclear norm. The related weighted nuclear norm proximal (WNNP) operator [51] with $F$-norm data fidelity is formulated as follows:

$$X = \text{prox}_{\frac{1}{\alpha}\|\cdot\|_{w,*}}(Y) = \arg\min_X \| X \|_{w,*} + \frac{\alpha}{2} \| Y - X \|_F^2 \tag{9}$$

Equation (9) is also referred to as the WNNM problem. WNNP operator is a generalization of the NNP operator given in equation (6). The WNNM problem is not convex in general case, and it is more difficult to solve than NNM problem. Literature [50] considered 3 cases of the arrangement of the weights. The first case is that the weights are in a non-ascending order. In this case the WNNM problem is still a convex optimization problem as the NNM problem and the problem has a globally optimal solution. The second case is that the weights are in an arbitrary order. In this case the WNNM problem is non-convex. The problem can be solved iteratively via sorting the diagonal elements and shrinking the singular values. The third case is that the weights are in a non-descending order. The iterative algorithm used to solve the problem in the second case will have a fixed point. Since the singular values with smaller magnitude mainly correspond to noise and errors in the data, these singular values corresponding to higher frequency of the data should be attenuated more. While the singular values with higher magnitude constitute most of the data variance, these singular values corresponding to lower frequency of the data should be attenuated less. Based on the above principle, literature [50] proposed to use the weights arranged in non-descending order for image denoising applications which penalizes less the higher singular values.

**Robust principal component analysis (RPCA).** While the above methods work well for data corrupted by Gaussian noise, they fail for data corrupted by gross errors. In [33] this issue is addressed by assuming that the outliers are sparse, i.e., only a small percentage of the entries of $D$ are corrupted. Hence, the goal is to decompose the data matrix $D$ as the sum of a low-rank matrix $A$ and a sparse matrix $E$, i.e.,

$$\min_{A,E} \text{rank}(A) + \gamma \| E \|_0 \text{ s.t. } D = A + E \tag{10}$$

where $\gamma > 0$ is a parameter. Since this problem is in general NP hard, a common practice is to replace the rank of $A$ by its nuclear norm and the $l_0$ norm by the $l_1$ norm. It is shown in [33] that, under broad conditions, the optimal solution to the problem in (10) is identical to that of the convex problem:

$$\min_{A,E} \| A \|_* + \gamma \| E \|_1 \text{ s.t. } D = A + E \tag{11}$$

While a closed form solution to this problem is not known, convex optimization techniques can be used to find the minimizer. Problem (11) is the well-known RPCA problem and it has wide applications in computer vision and pattern recognition.

**WNNM based RPCA.** WNNM can also be applied to the RPCA problem and the resulting WNNM-based RPCA (WNNM-RPCA) model can be written as [51]:

$$\min_{A,E} \| A \|_{w,*} + \gamma \| E \|_1 \text{ s.t. } D = A + E \tag{12}$$

This problem can also be solved by using convex optimization techniques. Besides WNNM-RPCA, literature [51] also proposed WNNM based matrix completion (WNNM-MC) which is to apply WNNM to the original NNM based matrix completion problem. WNNM-RPCA and WNNM-MC can be used in various low level vision applications including background subtraction and image inpainting. The weights are also arranged in non-descending order which penalizes less the higher singular values.

*2.2 WNN applied to the low rank representation model*

Inspired by the work done in the literature [50], [51], [52], in this paper, we propose to use WNN to regularize the representation matrix $Z$ in the low rank representation model (1). The WNNM based low rank representation model can be written as:

$$\min_{Z, E} \| Z \|_{w,*} + \lambda \| E \|_{2,1} \quad \text{s.t.} \quad X = XZ + E \tag{13}$$

Where the nuclear norm regularization of the representation matrix $Z$ is replaced by the WNN regularization. Incorporating the weights will improve the flexibility of the low rank representation model. Equation (13) is called WNNM-LRR model. As in NNM (nuclear norm minimization) -LRR, we also employ the ADMM to solve the WNNM-LRR problem. Equation (13) can be converted to the following equivalent problem:

$$\min_{Z, E, J} \| J \|_{w,*} + \lambda \| E \|_{2,1} \quad \text{s.t.} \quad X = XZ + E, Z = J \tag{14}$$

This problem can be solved via minimizing the following augmented Lagrangian function:

$$\mathcal{L} = \| J \|_{w,*} + \lambda \| E \|_{2,1} + \operatorname{tr}(Y_1^T (X - XZ - E)) + \operatorname{tr}(Y_2^T (Z - J)) + \frac{\mu}{2} (\| X - XZ - E \|_F^2 + \| Z - J \|_F^2)$$

Where $Y_1$ and $Y_2$ are Lagrange multipliers and $\mu > 0$ is a penalty parameter. The above problem is unconstrained. So it can be minimized with respect to $J$, $Z$ and $E$, respectively, by fixing the other variables and then updating the Lagrange multipliers $Y_1$ and $Y_2$. The optimization procedure is described in Algorithm 1.

---

**Algorithm 1** WNNM-LRR

---

**Input:** data matrix $X$, parameter $\lambda$, weight vector $w$

**Initialize:** $Z_{(0)} = J_{(0)} = 0$, $E_{(0)} = 0$, $Y_{1,(0)} = 0$, $Y_{2,(0)} = 0$, $\mu_{(0)} = 10^{-6}$, $\mu_{\max} = 10^6$, $\rho = 1.1$, $\varepsilon = 10^{-8}$ and $k=0$

**while** not converged **do**

    **1.** fix the others and update $J$ by

$$J_{(k+1)} = \arg\min_{J} \frac{1}{\mu_{(k)}} \| J \|_{w,*} + \frac{1}{2} \| J - (Z_{(k)} + Y_{2,(k)} / \mu_{(k)}) \|_F^2.$$

    **2.** fix the others and update $Z$ by

$$Z_{(k+1)} = (I + X^T X)^{-1}(X^T (X - E_{(k)}) + J_{(k+1)} + (X^T Y_{1,(k)} - Y_{2,(k)}) / \mu_{(k)}).$$

    **3.** fix the others and update $E$ by

$$E_{(k+1)} = \arg\min_{E} \frac{\lambda}{\mu_{(k)}} \| E \|_{2,1} + \frac{1}{2} \| E - (X - XZ_{(k+1)} + Y_{1,(k)} / \mu_{(k)}) \|_F^2.$$

    **4.** update the multipliers

$$Y_{1,(k+1)} = Y_{1,(k)} + \mu_{(k)} (X - XZ_{(k+1)} - E_{(k+1)}),$$

$$Y_{2,(k+1)} = Y_{2,(k)} + \mu_{(k)} (Z_{(k+1)} - J_{(k+1)}).$$

    **5.** update the parameter $\mu$ by $\mu_{(k+1)} = \min(\rho \mu_{(k)}, \mu_{\max})$.

    **6.** check the convergence conditions:

        $\| X - XZ_{(k+1)} - E_{(k+1)} \|_\infty < \varepsilon$ and $\| Z_{(k+1)} - J_{(k+1)} \|_\infty < \varepsilon$.

    **7.** $k=k+1$.

**end while**

**Output:** Matrix $Z = Z_{(k+1)}$ and $E = E_{(k+1)}$.

---

In step 1 of algorithm 1, a WNNM has to be solved. The weights adopted in this paper are in a non-ascending

order as opposed to the arrangement of the weights in previous works. The weights are chosen as:

$$w_i = \sigma_i^\gamma(X) \tag{15}$$

Where $\gamma$ is a parameter which determines the distribution of the weights. From equation (15) we can see that the weights are proportional to the singular value of the matrix to some power. Since the singular values are always in a non-ascending order, the weights will also be in a non-ascending order. Consequently, the optimization problem in step 1 is still a convex optimization problem and it has a closed form solution which can be written as:

$$J_{(k+1)} = U \mathcal{S}_{\frac{1}{\mu_{(k)}}w}(\Sigma) V^T$$

Where $U\Sigma V^T$ is the singular value decomposition of the matrix $Z_{(k)} + Y_{2,(k)}/\mu_{(k)}$.

Here we will give some intuitive explanation to equation (15). The singular values of the representation matrix contain important discriminative information of the data matrix. In the original low rank representation model, the smaller singular values are over thresholded while the larger singular values are inadequately thresholded by using the same soft-thresholding scheme. This will result the loss of the discriminative information in the representation matrix. In order to overcome this difficulty, we propose in this paper to penalize more the larger singular values and penalize less the smaller singular values in order to retain the discriminative information while simultaneously achieve the goal of nuclear norm minimization. The representation matrix thus obtained will have a small nuclear norm while the discriminative information will also be contained in the matrix.

## 3. Solving the convex problem via LADMM

Using conventional ADMM to solve the low rank representation model will have to replace the matrix $Z$ with the matrix $J$ in equation (1) and introduce another constraint $Z=J$ as can be seen in the equation (14). This can be explained as follows. When directly solving (1) by ADMM, the augmented Largrangian function is formulated as follows

$$\mathcal{L}'(Z,E,\mu') = \|Z\|_* + \lambda \|E\|_{2,1} + \mathrm{tr}(Y'(X-XZ-E)) + \frac{\mu'}{2}(\|X-XZ-E\|_F^2) \tag{16}$$

Then ADMM decomposes the minimization of $\mathcal{L}'$ with respect to $(Z, E)$ into two subproblems that minimize with respect to $Z$ and $E$ respectively. In particular, the iterations of ADMM go as follows:

$$Z_{(k+1)} = \arg\min_Z \mathcal{L}'(Z, E_{(k)}, \mu'_{(k)}) = \arg\min_Z \|Z\|_* + \frac{\mu'_{(k)}}{2}\|X - XZ - E_{(k)} + Y_{(k)}/\mu'_{(k)}\|_F^2 \tag{17}$$

$$E_{(k+1)} = \arg\min_E \mathcal{L}'(Z_{(k+1)}, E, \mu'_{(k)}) = \arg\min_E \lambda \|E\|_{2,1} + \frac{\mu'_{(k)}}{2}\|X - XZ_{(k+1)} - E + Y_{(k)}/\mu'_{(k)}\|_F^2 \tag{18}$$

$$Y'_{(k+1)} = Y'_{(k)} + \mu'_{(k)}(X - XZ_{(k+1)} - E_{(k+1)}) \tag{19}$$

Subproblem (18) has closed form solution but subproblem (17) does not since the linear mappings before $E$ is the identity matrix but the linear mappings before $Z$ is the matrix $X$ and it is not an identity matrix. In this case subproblem (18) has to be solved iteratively. To overcome this difficulty, a common strategy is to introduce an auxiliary variable $J$ and reformulate equation (1) into an equivalent one:

$$\min_{Z,E,J} \|J\|_* + \lambda \|E\|_{2,1} \quad \text{s.t.} \quad X = XZ + E, Z = J$$

Literature [55], [56] introduced LADMM to solve the subproblem when the linear mapping before the variable to be optimized is not an identity matrix. The method is to linearize the quadratic term in (17) at $Z_{(k)}$ and adding a proximal term and the approximation is

$$Z_{(k+1)} = \arg\min_{Z} \|Z\|_* + \text{tr}((X^T Y_{(k)} + \mu'_{(k)} X^T (X - XZ_{(k)} - E_{(k)}))^T (Z - Z_{(k)})) + \frac{\mu'_{(k)} \eta_X}{2} \|Z - Z_{(k)}\|_F^2$$

$$= \arg\min_{Z} \|Z\|_* + \frac{\mu'_{(k)} \eta_X}{2} \|Z - Z_{(k)} + X^T (Y_{(k)} + \mu'_{(k)} (X - XZ_{(k)} - E_{(k)})) / (\mu'_{(k)} \eta_X)\|_F^2$$

Where $\eta_X$ is a parameter. The above approximation resembles that of APG [57]. Using this formulation, $Z$ can be solved in closed form and the additional constraint $Z=J$ is not needed.

LADMM can improve the convergence performance of ADMM since no auxiliary variables are required. The LADMM proposed in the literature [55] is a general method and can be applied to other convex problems. The proposed WNNM-LRR model can also be solved using LADMM. The optimization procedure is presented in algorithm 2.

**Algorithm 2** WNNM-LRR solved via LADMM

**Input:** data matrix $X$, parameter $\lambda$, weight vector $w$

**Initialize:** $Z_{(0)} = 0$, $E_{(0)} = 0$, $Y' = 0$, $\mu'_{max} > \mu'_{(0)} > 0$, $\rho_0 > 1$, $\varepsilon_1 > 0, \varepsilon_2 > 0$, $\eta_X > \sigma_{max}^2(X)$, and $k=0$

**while** not converged **do**

  1. fix the others and update $E$ by

$$E_{(k+1)} = \arg\min_{E} \frac{\lambda}{\mu'_{(k)}} \|E\|_{2,1} + \frac{1}{2} \|E - (X - XZ_{(k)} + Y'_{(k)} / \mu'_{(k)})\|_F^2.$$

  2. fix the others and update $Z$ by

$$Z_{(k+1)} = \arg\min_{Z} \|Z\|_{w,*} + \frac{\mu'_{(k)} \eta_X}{2} \|Z - Z_{(k)} + X^T (Y_{(k)} + \mu'_{(k)} (X - XZ_{(k)} - E_{(k+1)})) / (\mu'_{(k)} \eta_X)\|_F^2.$$

  3. update the multipliers

$$Y'_{(k+1)} = Y'_{(k)} + \mu'_{(k)} (X - XZ_{(k+1)} - E_{(k+1)}).$$

  4. update the parameter $\mu'$ by $\mu'_{(k+1)} = \min(\rho' \mu'_{(k)}, \mu'_{max})$. The value of $\rho'$ is defined as

$$\rho' = \begin{cases} \rho_0, & \text{if } \mu'_{(k)} \max(\|Z_{(k+1)} - Z_{(k)}\|_F, \|E_{(k+1)} - E_{(k)}\|_F) / \|X\|_F < \varepsilon_2 \\ 1, & \text{otherwise} \end{cases}$$

  5. check the convergence conditions:

   $\|X - XZ_{(k+1)} - E_{(k+1)}\|_F / \|X\| < \varepsilon_1$ and

   $\max(\|Z_{(k+1)} - Z_{(k)}\|_F, \|E_{(k+1)} - E_{(k)}\|_F) / \|X\|_F < \varepsilon_2$.

  6. $k=k+1$.

**end while**

**Output:** Matrix $Z = Z_{(k+1)}$ and $E = E_{(k+1)}$.

The WNNM-LRR solved via LADMM is referred to as WNNM-LRR(L) for short. Using LADMM, the WNNM based low rank representation model can be optimized very efficiently.

## 4. Experimental results and analysis

In this section, we evaluate the performance of the proposed WNNM-LRR and WNNM-LRR(L) in dealing with two real-world problems: clustering images of human faces and clustering images of hand written digits. The performance of the proposed method is compared with the best state-of-the-art subspace clustering algorithms: LSA [25], SLBF [27], LLMC [26], SCC [29], MSL [17], LRR [39] and SSC [36]. LSA, SLBF, LLMC, SCC, LRR and SSC are all spectral clustering-based algorithms where LSA, SLBF, LLMC are local spectral clustering-based algorithms and SCC, LRR and SSC are global spectral clustering-based algorithms. MSL can be considered as a kind of iterative statistical method.

**Implementation details.** In WNNM-LRR and WNNM-LRR(L), the weighting parameter $\lambda$ is chosen as $1/\sqrt{\log(n)}$ as suggested by the literature [58], the parameter $\gamma$ which determines the distribution of the weights is chosen as 1/3. For the comparison algorithms, we use the codes provided by their authors. For LSA, we use $K$=15 nearest neighbors and dimension $d$=9, to fit local subspaces for face clustering and use $K$=5 nearest neighbors and dimension $d$=9 for handwritten digit clustering. For SLBF, we use dimension $d$=9 to fit local subspaces for face clustering and handwritten digit clustering. The start size and step size of local neighborhood selection is respectively set as 10 and 5 for face clustering and 2 and 2 for handwritten digit clustering. For LLMC, we use $K$=10 nearest neighbors for face clustering and $K$=5 nearest neighbors for handwritten digit clustering. For SCC, we use dimension $d$=9 to fit local subspaces and use the affine SCC clustering version for face clustering and handwritten digit clustering. For MSL, the initial segmentation is provided by the shape interaction matrix method and dimension $d$=9 is chosen to fit local subspaces for face clustering and handwritten digit clustering. For LRR, we also use the suggested weighting parameter $\lambda = 1/\sqrt{\log(n)}$. For SSC, the optimization program with affine constraint and noise term and sparse outlying term is used with the weighting parameter $\lambda_z = 800/\mu_z$ for the noise term and $\lambda_e = 20/\mu_e$ for the sparse outlying term as suggested by the authors. Finally, as LSA, SLBF, LLMC, SCC and MSL need to know the number of subspaces a prior and the estimation of the number of subspaces from the eigenspectrum of the graph Laplacian in the noisy setting is often unreliable [36], to have a fair comparison we provide the number of subspaces as an input to all the algorithms.

**Datasets.** For the face clustering problem, we consider the Extended Yale B dataset [49], which consists of face images of 38 human subjects, where images of each subject lie in a low-dimensional subspace [3]. For the handwritten digit clustering problem, we consider the MNIST dataset and USPS dataset. The MNIST dataset of handwritten digits has a training set of 60000 examples and a test set of 10000 examples. The digits have been size-normalized and centered in a fixed-size image. The USPS dataset consists of a training set with 7291 images and a test set with 2007 images. The digits have also been size-normalized. The images of each digit also lie in a low-dimensional subspace. The details of all datasets used are provided in Table 2.

Table 2 Details of the datasets used for clustering experiments in this work

| Dataset | Training samples | Test samples | Dimension | Classes |
|---|---|---|---|---|
| Extended Yale B | / | 2432 | $192 \times 168$ | 38 |
| MNIST | 60000 | 10000 | $28 \times 28$ | 10 |
| USPS | 7291 | 2007 | $16 \times 16$ | 10 |

*4.1 Face clustering*

The face images in the Extended Yale B dataset are acquired under a fixed pose with varying lighting conditions. It has been shown that, under the Lambertian assumption, images of a subject with a fixed pose and vary lighting conditions lie close to a linear subspace of dimension 9 [3], [36]. Thus, the collection of face images of multiple subjects lies close to a union of 9D subspaces [36], [39].

In this section, the clustering performance of WNNM-LRR and WNNM-LRR(L) as well as the state-of-the-art methods on the Extended Yale B dataset [49] are evaluated and compared. The dataset consists of

$192\times168$ pixel cropped face images of $n=38$ individuals, where there are $N_i = 64$ frontal face images for each subject acquired under various lighting conditions. As done in the literature [36], we also downsample the images to $48\times42$ pixels to reduce the computational cost and the memory requirements of all algorithms. Thus each data point is a 2016D vector, hence $m=2016$.

We divide the 38 subjects into four groups as done in [36], where the first three groups correspond to subjects 1 to 10, 11 to 20, 21 to 30, and the fourth groups corresponds to subjects 31 to 38. The four groups are referred to as group 1-4 respectively. The authors in [36] consider all choices of $n \in \{2,3,5,8,10\}$ subjects for each of the first three groups and $n \in \{2,3,5,8\}$ for the last group. In this paper, we only consider $n=10$ for each of the first three groups and $n=8$ for the last group since this is the most challenge situation for face clustering. In [36], the author reported clustering results in three different settings: the first is applying RPCA separately on each subject and then use the clustering algorithm, the second is applying RPCA simultaneously on all subject and then use the clustering algorithm and the third is directly using the clustering algorithm to the original data points. In this paper, the first and second settings are not considered and all the clustering algorithms are directly applied to the original data points. The results are shown in Table 3 from which we make the following conclusions:

Table 3 Clustering accuracy (%) of different algorithms on the Extended Yale B dataset without preprocessing the data

| Algorithm | LSA[25] | SLBF[27] | LLMC[26] | SCC[29] | MSL[17] | SSC[36] | LRR[39] | WNNM-LRR | WNNM-LRR(L) |
|---|---|---|---|---|---|---|---|---|---|
| Group 1 | 25.00 | 37.26 | 40.09 | 29.25 | 64.94 | 83.18 | 95.44 | 96.54 | 81.45 |
| Group 2 | 25.68 | 38.36 | 43.18 | 20.39 | 59.55 | 89.89 | 85.23 | 84.91 | 97.59 |
| Group 3 | 25.31 | 34.06 | 32.34 | 30.31 | 71.41 | 94.22 | 96.72 | 97.34 | 98.13 |
| Group 4 | 30.44 | 40.12 | 38.54 | 30.63 | 42.69 | 78.06 | 90.32 | 93.08 | 95.46 |

Local spectral clustering-based algorithms obtain a relatively low clustering accuracy on the dataset. Local spectral clustering-based algorithms use the K nearest neighbors of a point to estimate a local linear subspace of that point. As for the Extended Yale B dataset, there are a relatively large number of points near the intersection of subspaces; the local neighborhood of a point must contain a lot of points from other subspaces. Thus the local linear subspace estimated is not accurate and thus the similarity built based on this local subspace is not accurate either.

Global spectral clustering-based algorithm such as SCC also obtains low clustering accuracy. Although global spectral clustering-based algorithm doesn't rely on the nearest neighbors, the noises and outliers exist in the dataset will affect the estimation of the polar curvature and thus similarity built based on this polar curvature cannot reflect the actual similarity between the data points.

Global spectral clustering-based algorithms such as LRR and SSC obtain more accurate clustering results while LRR gets higher clustering accuracy than SSC. LRR and SSC both exploit the global structure of the dataset and handle noises and outliers explicitly. The representation matrix obtained can reflect the relationship of the data point with other points and the similarity matrix is accurate.

Iterative statistical method such as MSL obtains relatively low clustering accuracy. This method needs an initial clustering of the dataset. This method then uses an iterate method to refine the initial clustering results. This method can easily get stuck in local minimum around the initial segmentation and the clustering accuracy is thus highly dependent on the initial clustering. Here the shape interaction matrix method is used as the initial clustering method. Since the rank of the data matrix is not known beforehand, the clustering accuracy is relatively low based on the estimated rank of the data matrix.

WNNM-LRR obtains higher clustering accuracy as compared to LRR and SSC. Here we want to make a comparison of the distribution of the singular values of the representation matrix obtained by LRR and WNNM-LRR. The distribution of the singular values of the representation matrix obtained using the dataset of group 1 is shown in Fig. 1. From the figure, we can find that there are more small singular values of WNNM-LRR and the transition of the large singular values to the smaller ones are more smooth. Since the weights used in our method are arranged in a non-ascending order, larger singular values are penalized more while smaller singular values are penalized less and thus smaller singular values are reserved in the process of the singular value thresholding. In contrast, in the singular value thresholding procedure of LRR, all singular values are thresholded with the same extent causing the loss of smaller singular values. Since smaller singular values also contain the information about the data structures, they should be reserved.

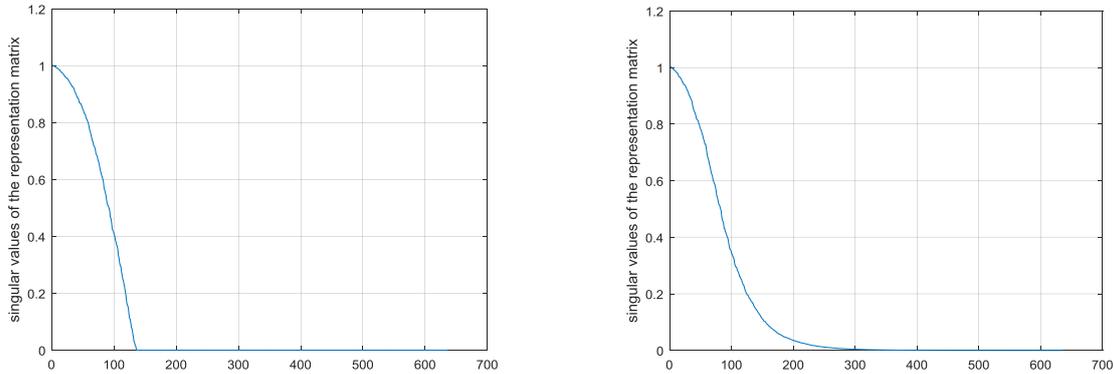

(a) the singular values obtained by LRR     (b) the singular values obtained by WNNM-LRR

Figure 1 The distribution of the singular values of the representation matrix obtained by LRR and WNNM-LRR

As for WNNM-LRR(L), we can see the clustering accuracy is further improved on groups 2-4. When using the linearized technique, the representation matrix corresponding to the variable $Z$ is directly solved in closed form. In contrast, in original ADMM, the variable $Z$ must be replaced by another variable and an extra equality constraint is introduced. Directly solving the variable $Z$ can lead to more accurate estimation of the representation matrix and using the weighted nuclear norm can reserve the small singular values, thus lead to higher clustering accuracy.

Here we also want to point out an implementation detail in realizing the spectral clustering-based algorithms. In spectral clustering-based algorithms, a similarity matrix is formed first and then spectral clustering is applied to the similarity matrix. In the last procedure of spectral clustering, k-means clustering algorithm is used to cluster the eigenvectors of the Laplacian matrix. In the program provided by the authors, the k-means clustering algorithm is initialized with random clustering centers and multiple restarts are used. As pointed by [23], the clustering centers can be selected in a deterministic way. The clustering centers of the k-means algorithm used in the algorithm SCC is also initialized with a deterministic way. We adopt this way of initializing clustering centers of the k-means algorithm. Through experiments we found that using deterministic way of initializing clustering centers can get higher clustering accuracy as compared to the random initialization with multiple restarts.

*4.2 Handwritten digit clustering*

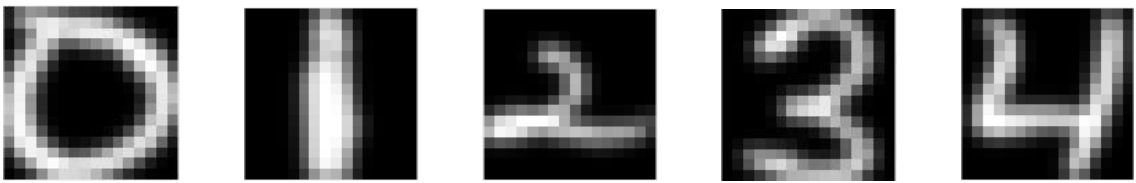

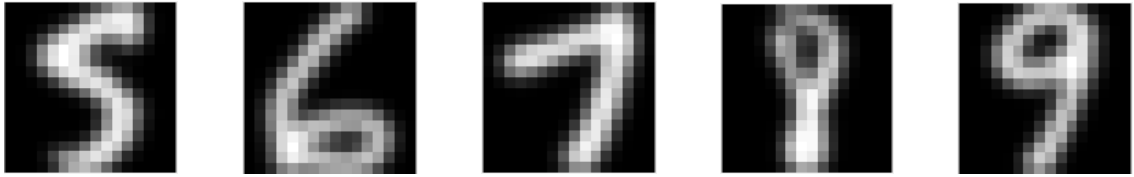

Figure 2 Some data samples from the USPS dataset

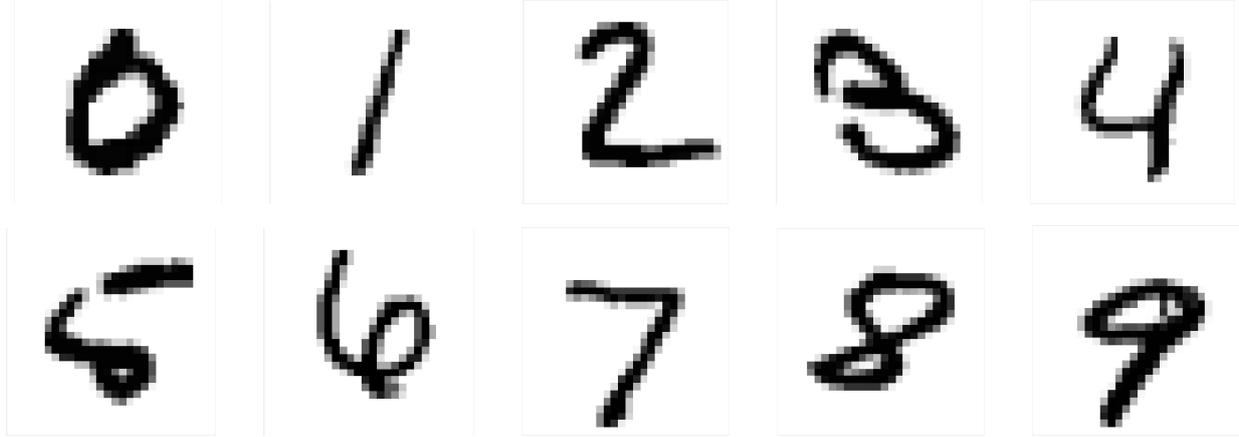

Figure 3 Some data samples from the MNIST dataset

Handwritten digit clustering refers to the problem of clustering multiple handwritten digits with different rotations and translations. We use the handwritten digits from the USPS dataset and MNIST dataset to evaluate the performance of different subspace clustering algorithms. Some data samples from the USPS dataset and MNIST dataset are shown in Fig.2 and Fig.3 respectively. The handwritten digits are hard to cluster for the following reasons. Frist, there are relatively less information contained in each image because each image can be seen as a binary image with black background and white strokes or white background and black strokes. Second, as we can see from the figure, the distance between the subspaces are close to each other because some numbers are similar to others, for example, number 1 is similar to number 7, number 2 is similar to number 5 and number 6, 8, 9 are all similar with each other. It is difficult to distinguish between these numbers using subspace clustering algorithms. We take all the 10 classes and each class contains 10 samples as the data matrix for subspace clustering. For each database, we take numbers from the training samples and testing samples respectively and they are referred to as train group and test group. The clustering accuracy for the USPS dataset and MNIST dataset is presented in Table 4 and Table 5 respectively.

Table 4 Clustering accuracy (%) of different algorithms on the USPS dataset

| Algorithm | LSA[25] | SLBF[27] | LLMC[26] | SCC[29] | MSL[17] | SSC[36] | LRR[39] | WNNM-LRR | WNNM-LRR(L) |
|---|---|---|---|---|---|---|---|---|---|
| Train group | 63.00 | 52.00 | 32.00 | 37.00 | 65.00 | 34.00 | 75.00 | 77.00 | 66.00 |
| Test group | 69.00 | 53.00 | 27.00 | 38.00 | 65.00 | 34.00 | 56.00 | 61.00 | 59.00 |

Table 5 Clustering accuracy (%) of different algorithms on the MNIST dataset

| Algorithm | LSA[25] | SLBF[27] | LLMC[26] | SCC[29] | MSL[17] | SSC[36] | LRR[39] | WNNM-LRR | WNNM-LRR(L) |
|---|---|---|---|---|---|---|---|---|---|
| Train group | 55.00 | 54.00 | 33.00 | 33.00 | 61.00 | 60.00 | 24.00 | 55.00 | 52.00 |
| Test group | 58.00 | 42.00 | 47.00 | 28.00 | 60.00 | 57.00 | 36.00 | 55.00 | 54.00 |

From the results, we make the following conclusions: all the subspace clustering algorithms get relatively low clustering accuracy as compared to the results on the Extended Yale B dataset. The proposed WNNM-LRR and WNNM-LRR(L) algorithm is better than the original LRR algorithm. In the USPS dataset, WNNM-LRR get the highest clustering accuracy on the train group, LSA get the highest clustering accuracy on the test group. In the MNIST dataset, MSL get the highest clustering accuracy on both the train group and test group. Here we give an explanation of the low clustering accuracy of the original LRR algorithm. There are two main reasons: first, since the numbers are similar to each other, the LRR algorithm will tend to cluster the numbers with similar shape into a single cluster; second, since the noises and outliers in the handwritten digits cannot be well described by the sample-specific error, the original LRR which uses sample-specific error is not appropriate in clustering hand written digits.

## 5. Conclusions and future work

We studied the problem of clustering a collection of data points that lie in or close to a union of low-dimensional subspaces. We proposed a subspace clustering algorithm based on low rank representation and weighted nuclear norm minimization, with original ADMM and linearized ADMM optimization method. The method first finds a representation matrix and during the singular value thresholding procedure, different weights are used with different singular values and smaller singular values are reserved. Experiments on real data set such as face images and handwritten digits show the effectiveness of our algorithm and its superiority over the state of the art.

Here we also point out several possible future research directions. Since small singular values of the representation matrix are useful for clustering, other methods which can reserve the small singular values will improve the clustering accuracy as expected. In the future research, we will find other methods to reserve the small singular values. In original LRR, the data matrix itself is used as the dictionary. Since the data matrix is contaminated with noises and outliers, directly using the data matrix as the dictionary is not appropriate. In a recent research work [43], the author suggested using a clean dictionary to represent the data matrix. The weighted nuclear norm minimization technique can also be introduced to this method. Finally, all the existing improved LRR methods which use rank minimization can be further improved using the weighted nuclear norm minimization technique.

*Acknowledgment*

This work is partially supported by the Key Laboratory of Port, Waterway and Sedimentation Engineering of the Ministry of Transport, Nanjing Hydraulic Research Institute; State Key Laboratory of Urban Water Resource and Environment under Grant ES201409; and Key Laboratory of Rivers and Lakes Governance and Flood Protection of Yangtse River Water Conservancy Committee under Grant CKWV2013225/KY.